\batchmode
\makeatletter
\def\input@path{{D:/Downloads/IV2018/}}
\makeatother
\documentclass[twocolumn,english,conference]{IEEEtran}
\usepackage[T1]{fontenc}
\usepackage[latin9]{inputenc}
\usepackage{color}
\usepackage{verbatim}
\usepackage{amsmath}
\usepackage{amssymb}
\usepackage{graphicx}

\makeatletter

\providecommand{\tabularnewline}{\\}

\usepackage{cite}
\usepackage[margin=0.75in]{geometry}
\usepackage{amsmath}

\usepackage{babel}
\usepackage{subfig}
\usepackage{caption}

\makeatother

\usepackage{babel}
\begin{document}

\title{\vspace{0.25in}Road Segmentation Using CNN with GRU}

\author{Yecheng Lyu and Xinming Huang\\
 Department of Electrical and Computer Engineering\\
Worcester Polytechnic Institute\\
Worcester, MA 01609, USA\\
\{ylyu,xhuang\}@wpi.edu}
\maketitle
\begin{abstract}
This paper presents an accurate and fast algorithm for road segmentation
using convolutional neural network (CNN) and gated recurrent units
(GRU). For autonomous vehicles, road segmentation is a fundamental
task that can provide the drivable area for path planning. The existing
deep neural network based segmentation algorithms usually take a very
deep encoder-decoder structure to fuse pixels, which requires heavy
computations, large memory and long processing time. Hereby, a CNN-GRU
network model is proposed and trained to perform road segmentation
using data captured by the front camera of a vehicle. GRU network
obtains a long spatial sequence with lower computational complexity,
comparing to traditional encoder-decoder architecture. The proposed
road detector is evaluated on the KITTI road benchmark and achieves
high accuracy for road segmentation at real-time processing speed.
\end{abstract}

\begin{IEEEkeywords}
autonomous vehicle, road segmentation, gated recurrent units
\end{IEEEkeywords}

\section{Introduction}

In recent years, there is a growing research interest on automated
driving and intelligent vehicles. As one of the most important parts
in an automated driving system, road perception algorithm first gathers
information from the road and sets up the constraints for the subsequent
path planners \cite{hillel2014RoadDetection}\cite{son2015IlluminationInvariantLaneDetection}\cite{chen2017LidarHisto}\cite{khosroshahi2016surround_vehicle_traj_RNN}\cite{Qi2016HDT}.
Then it searches for the drivable area and the lane occupancy so that
the region of path planning and lane keeping can be determined. In
this paper we focus on the road segmentation algorithm using a monocular
camera input.

Cameras are the most popular sensors for autonomous and intelligent
vehicles since they are cost effective. There are existing test benches
such as KITTI\cite{fritsch2013KITTI_road} providing annotated images
for the evaluation of road/lane segmentation. Traditional computer
vision based road segmentation algorithms often employ manually defined
features such as edge \cite{yoo2013IlluminationRobustLaneDetection}
and histogram\cite{son2015IlluminationInvariantLaneDetection}. However,
manually defined features usually work on limited problem aspects
and hard to be extended to new domains\cite{hillel2014RoadDetection}.
Since 2014, CNN based deep learning algorithms have become more popular.
CNN is a kind of neural network that takes the advantage of many parallel
and cascade convolutional filters to solve high-dimensional non-convex
problem such as regression, image classification, object detection
and semantic segmentation. By processing limited dimensions and sharing
weights in each layer, a CNN requires fewer parameters than the traditional
artificial neural network and is much easier to train. From AlexNet\cite{NIPS2012_4824},
GoogleNet\cite{DBLP:journals/corr/SzegedyLJSRAEVR14}, VGGNet \cite{DBLP:journals/corr/SimonyanZ14a},
InceptionNet-v3 \cite{szegedy2016rethinking} to ResNet \cite{he2016deep}
, convolutional neural networks are growing larger that results better
performance. Several famous convolutional neural networks are compared
in accuracy and efficiency as shown in Table \ref{Table: Comparation of parameters and accuracy between nerual netowrks}.
\cite{Canziani2016CNN_review} also shows a detailed comparison among
different CNNs.

By implementing deeper layers and trainable parameters, convolutional
neural networks achieve amazing performance in variant light conditions,
scales and shapes. Unfortunately as networks become deeper and larger,
they take more computation, memory and processing time, which exceeds
the capability of the embedded systems in an autonomous vehicle. In
2017, a few efficient CNNs are introduced to lower down the parameters
and computational complexity for embedded devices. SequeezeNet \cite{Iandola2017SqueezeNet},
MobileNet \cite{DBLP:journals/corr/HowardZCKWWAA17}, ShuffleNet \cite{DBLP:journals/corr/ZhangZLS17}
and Xception \cite{DBLP:journals/corr/Chollet16a} are the state-of-art
efficient CNNs that separate pixel wise convolutions and dense wise
convolutions by applying grouped convolution and $1\times1$ convolutions.
Those efficient CNNs achieved competitive accuracy with less memories
and processing time if compared to the traditional CNNs. 

Recurrent neural network (RNN) is a kind of neural network structure
that passes data sequence. Different from traditional artificial neural
networks that fully connect all nodes and convolutional neural networks
that explore nodes from local to global layer by layer, recurrent
neural networks use state neurons to explore the relationship in context.
Simple RNN, LSTM \cite{article} and GRU \cite{DBLP:journals/corr/ChungGCB14}
are typical recurrent neural networks. RNNs have been proposed to
solve hard sequence problems such as machine translation\cite{Kalchbrenner2013},
video caption\cite{Venugopalan2015}. Most recently, RNNs have also
been used to solve spatial sequence \cite{NIPS2015_5955}, 2D image
sequence \cite{Oord2016Pixel_RNN} and spatial-temporal sequence \cite{DBLP:journals/corr/abs-1709-04875}
problems.

In this paper, the problem of road segmentation is framed as a semantic
caption task. The top, left and right boundaries of road area in an
image are extracted by a CNN-GRU network. A CNN based local feature
extractor and a GRU based context processor is implemented to construct
the network. The proposed solution is trained and evaluated on KITTI
road benchmarks and the results are satisfactory. We claim the proposed
network is embedded system friendly and is ready for real-time applications.
The rest of paper is organized as follows. Section \ref{sec:Algorithms-Design}
describes the proposed architecture. In section \ref{sec:Experiment}
experimental results on the benchmarks are presented and analyzed.
Finally Section \ref{sec:Conclusion} concludes the paper.

\begin{table*}
\centering{}%
\begin{tabular}{|c|c|c|c|c|}
\hline 
Network &
Publish year &
Parameters &
Multi-Adds &
Top-1 accuracy on ImageNet\tabularnewline
\hline 
\hline 
AlexNet\cite{NIPS2012_4824} &
2012 &
60M &
666M &
55\%\tabularnewline
\hline 
GoogleNet\cite{DBLP:journals/corr/SzegedyLJSRAEVR14} &
2014 &
6.8M &
1.43G &
68\%\tabularnewline
\hline 
VGG-16t\cite{DBLP:journals/corr/SimonyanZ14a} &
2014 &
138M &
15.3G &
72\%\tabularnewline
\hline 
InceptionNet-v3\cite{szegedy2016rethinking} &
2016 &
23M &
5.72B &
78\%\tabularnewline
\hline 
ResNet-101\cite{he2016deep} &
2016 &
44.5M &
31.5B &
77\%\tabularnewline
\hline 
\end{tabular}\caption{A comparison of parameters and accuracy among neural networks}
\label{Table: Comparation of parameters and accuracy between nerual netowrks}
\end{table*}

\section{Algorithms Design\label{sec:Algorithms-Design}}

In this section, the overview of the proposed neural network is described,
followed by the details of the main components including coordinate
input channels, local feature encoder and context processor.

\subsection{Coordinate Input Channels}

\begin{figure}
\includegraphics[width=0.9\columnwidth]{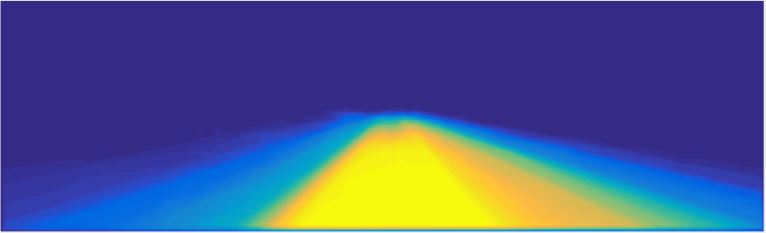}

\caption{Road area heat map on KITTI dataset}

\label{Fig: Road Heatmap}
\end{figure}

Coordinate of pixels in a road image is important to perception tasks.
Most of the existing CNN based detection and segmentation solutions
are trained with a large collection of images such as ImageNet, in
which coordinates of pixels/cells are not taken as features because
the camera views vary from image to image and objects may exist anywhere
in an image. For road perception, however, there is a strong coherence
between the likelihood of existence, shape and pose of an object and
its position in an image. For example, road pavement is more likely
located at the bottom of an image captured by the front view camera,
while cars have more chances to be smaller in the center of an image
and larger on the side, etc. Figure \ref{Fig: Road Heatmap} shows
the traffic scene heat map by analyzing KITTI dataset \cite{fritsch2013KITTI_road}
that indicates the possibility of a pixel belonging to the road area.
The closer a pixel locates to the center horizontally and bottom vertically,
the more likely it denotes to road area. Several research works have
taken coordinate input into consideration. In YOLO \cite{redmon2016yolo9000}
and MultiNet \cite{teichmann2016multinet}, images are divided into
cells and a convolutional neural network is built to process all cells
in one image. Coordinates are involved in the convolution of cells.
In \cite{brust2015efficient} and \cite{chen2017rbnet}, coordinates
are introduced at the end of CNN structure as bias on decoders. In
our solution, coordinates are introduced directly along with color
channels to provide more position related information as we did in
our previous work \cite{yecheng2017Road_LIDAR}.

\subsection{Local Feature Encoder}

\begin{figure}
\includegraphics[width=0.95\columnwidth]{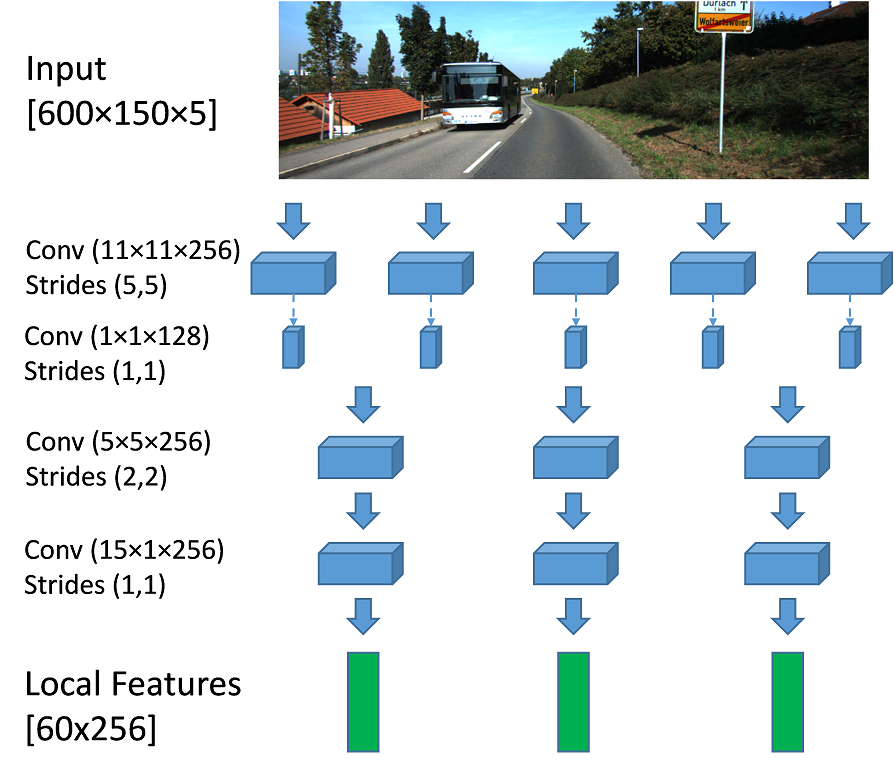}\caption{Network diagram of the local feature encoder}
\label{Fig: Network Description Diagram: Local Feature Extractor}

\end{figure}

Local feature encoder is a CNN based network that extracts features
such as illumination and edges from local patches. In CNNs, local
features are usually extracted by a group of convolution kernels trained
by a large number of samples for a specific task. Traditional CNN
based encoders such as FCN \cite{long2015fully} cascades a number
of convolution layers and each convolution layer grouped with pooling
and non-linear functions, which requires large memories and extensive
computations. In the proposed network, we implement a shallow structure
with large kernels followed by $1\times1$ convolutional layer as
is used in \cite{DBLP:journals/corr/ZhangZLS17} and \cite{DBLP:journals/corr/Chollet16a}.
The first convolution layer in the encoder is constructed by shallow
$11\times11\times256$ convolution kernels instead of using four convolution
layers as in FCN \cite{long2015fully}. Figure \ref{Fig: Network Description Diagram: Local Feature Extractor}
presents the detailed structure of the proposed encoder. To generate
enough features while limiting the computational complexity, we implement
$1\times1\times128$ convolution to reduce the dimension. Subsequently,
another $5\times5\times256$ and a $15\times1\times256$ convolution
layer is applied to further encode local features. Finally we encode
the $600\times150\times5$ input into $60$ feature vectors and each
vector contains $256$ features. 

\subsection{Context Processor}

Beside local features, context information throughout the entire image
is also important to road segmentation. CNN based solutions use a
deep encoder-decoder structure passing through all tensors in the
feature map to achieve context processing. This kind of structures
usually require very large GPU memories, vast amount of floating-point
operations and long processing time. However, embedded systems in
an intelligent vehicle have limited computational resources but require
real-time processing speed. In StixelNet \cite{levi2015stixelnet},
conditional random field (CRF) is applied for context processing.
It saves memory and float operations significantly with the penalty
of slow processing speed, approximately at 1 second per frame. In
our work, a GRU network is applied as context processor since it not
only has rich gates to handle diverse features but also is capable
of training as end-to-end. In our work, columns of feature vector
are queued to GRU and context information is stored in its hidden
states. Since rich context information is contained in both directions
(from left to right and from right to left), a bi-directional GRU
is build to process feature vector sequences in both directions. An
implementation of 128 neurons is set as hidden state for each direction,
so each context processor contains 256 neurons in total. 

\begin{figure}
\includegraphics[width=0.95\columnwidth]{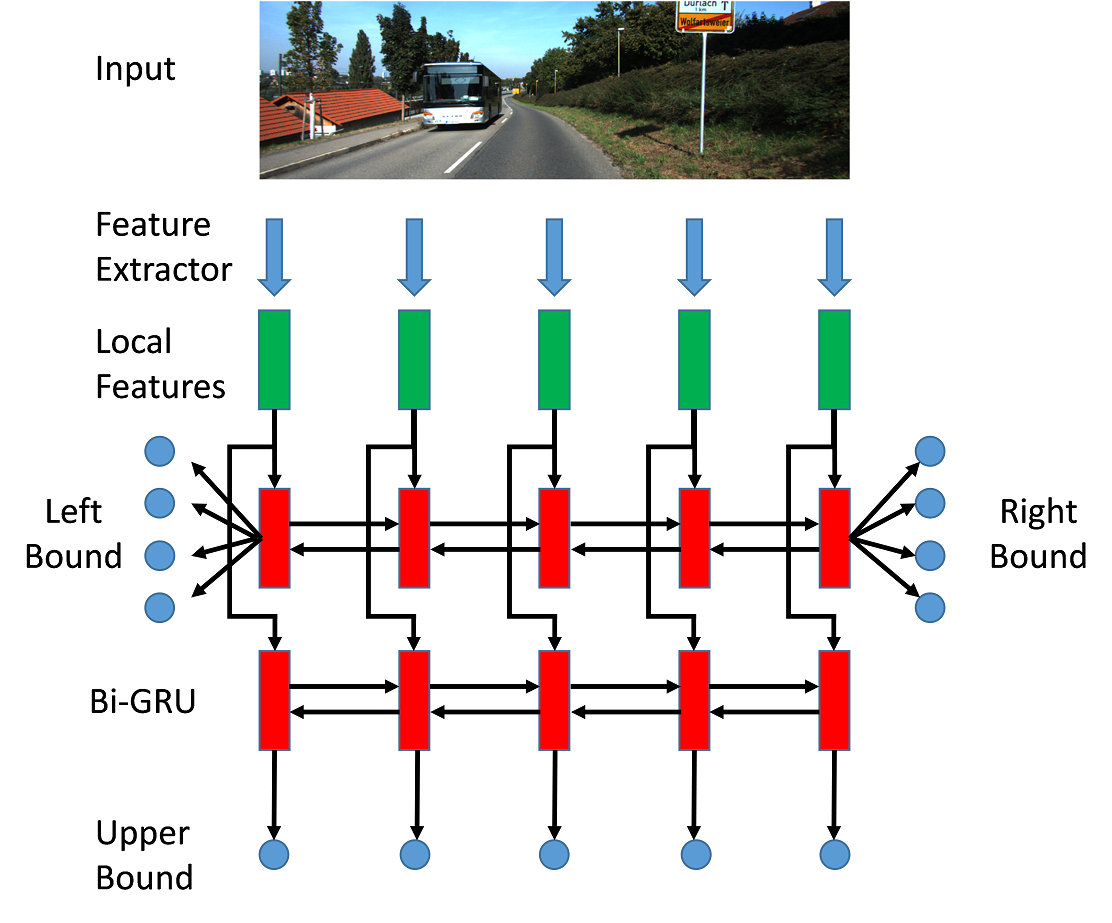}

\caption{Network diagram of the semantic processors}

\label{Fig: Network Diagram}
\end{figure}

In our method, two context processors are built. As is shown in Figure
\ref{Fig: Network Diagram}, the first processor is built to predict
the left and right bounds of road area and returns an output vector
after it processes the full sequence of feature vectors. The second
processor is used to predict the upper boundary of road area and returns
an output vector every time it processes a feature vector. Both context
processors are followed by a two-layer decoder to interpret the vector
into normalized position of boundaries. An up sampling layer is applied
to match the input height and width. The left, right and upper boundaries
separate non-road pixels in the image. Combining with the bottom of
an image as the default lower boundary, the total contoured area is
marked as the road area as the road segmentation output.

\subsection{Network Structure}

\begin{figure}
\includegraphics[width=0.95\columnwidth]{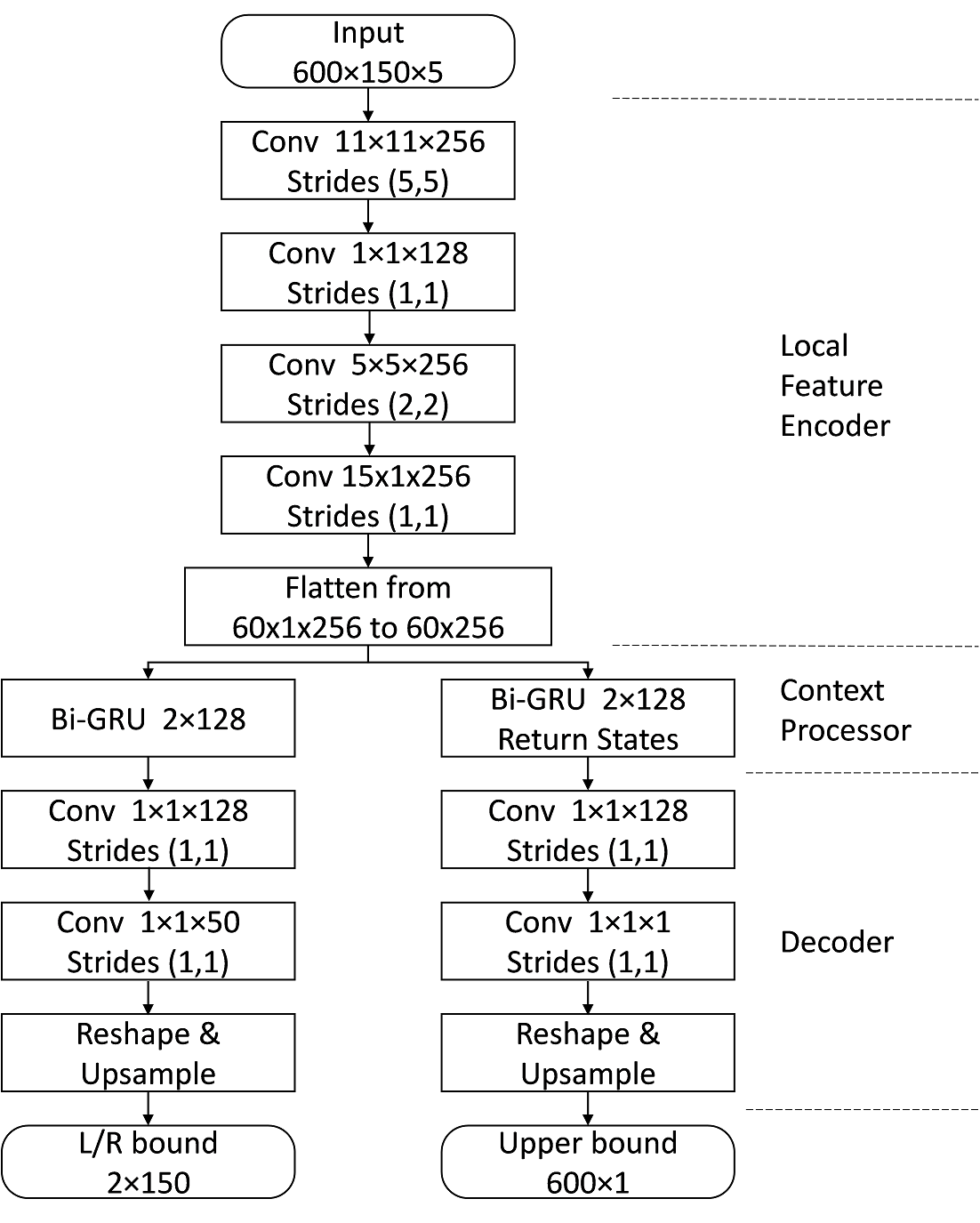}

\caption{Detailed architecture of the proposed CNN-RNN network }
\label{Fig: Detailed Network Body Architecture}

\end{figure}

The overall network architecture is shown in Figure \ref{Fig: Detailed Network Body Architecture}.
Input of the neural network is $600\times150\times5$. The first three
input channels are red, green and blue channels coming from camera
data, augmented with two additional channels as the row and column
coordinate of each pixel. In order to converge in training, all RGB
channels are divided by 255, row channel is normalized by image height
and column channel is divided by image width so that all input channels
are normalized to $[0,1]$ range. By passing through the local feature
encoder, context processors and decoders, the left, right and upper
boundaries of the road area in the image can be generated. For better
convergence in training session, the output boundaries are normalized
to $[0,0.5]$ range. 

For evaluation and visualization purpose, a binary map is generated
according to the output boundary. The predicted road area is the pixels
enclosed by those three boundaries and the bottom of the image.

\subsection{Pyramid Prediction Scheme}

\begin{figure}
\includegraphics[width=0.95\columnwidth]{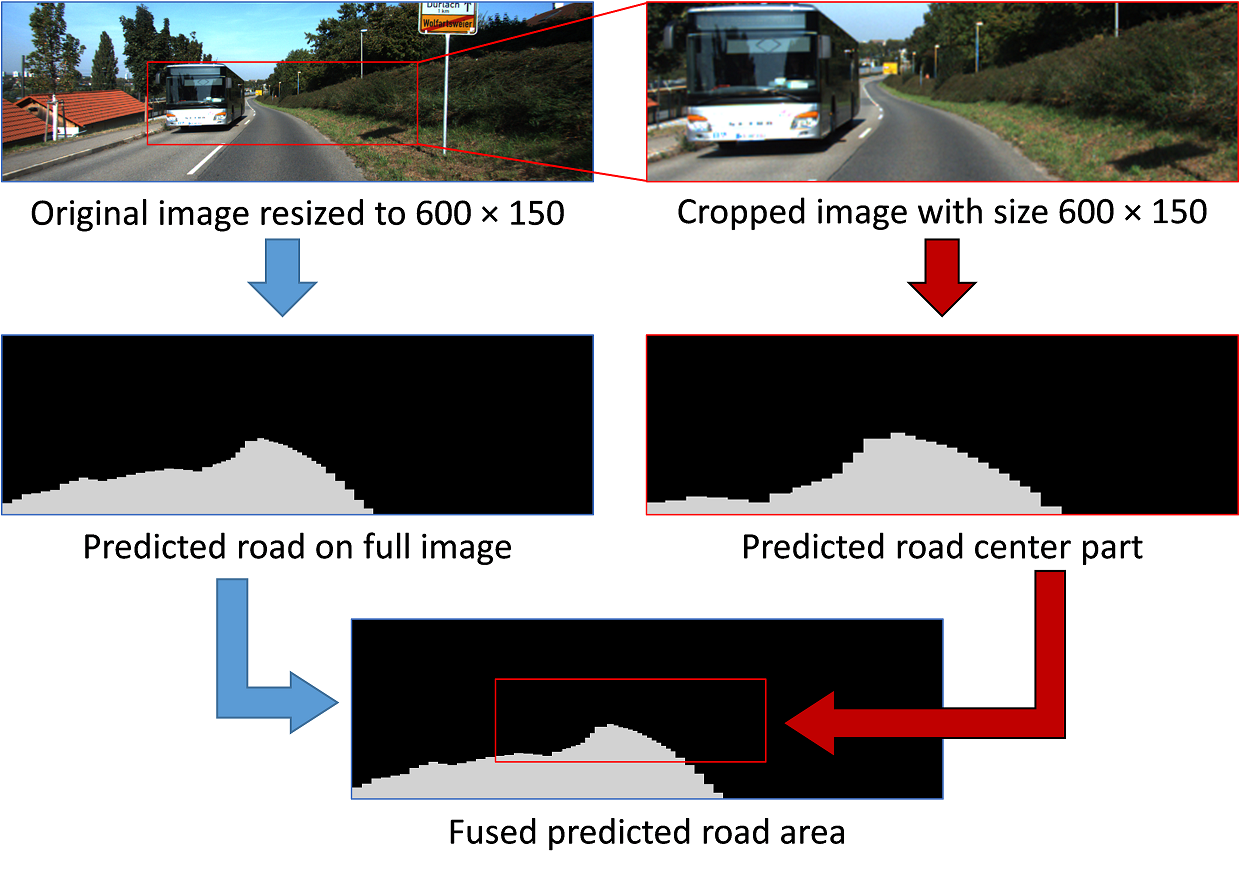}\caption{An illustration of the pyramid prediction scheme}
\label{Fig: Pyramid Prediction}

\end{figure}

Within each image frame, the features in near range and far range
are dramatically different in size. Simply scaling an entire image
frame to fit the input size of the network would result an unacceptable
level of feature loss in far range and cause low accuracy of detecting
the road further in distance. To avoid this problem, we propose a
pyramid prediction scheme. When predicting the road area in the near
range, the image frame is scaled to $600\times150$ before sending
to the network. When predicting the road area in the far range, the
image frame is cropped to $600\times150$ to match the network input
size. By applying the pyramid prediction scheme, road area in near
range and far range are predicted separately so that features in both
ranges can be scaled to similar size, which makes our local feature
encoder more stable and easier to train.

\section{Experiment \label{sec:Experiment}}

\begin{figure}
\includegraphics[width=0.95\columnwidth]{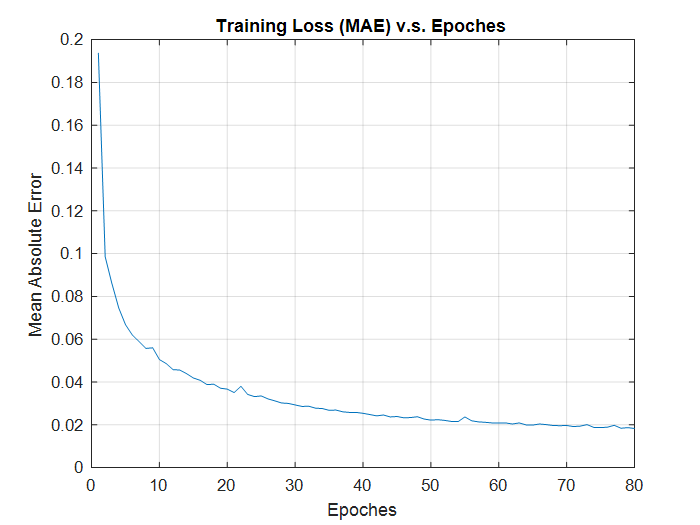}

\caption{Training loss measured by MAE }
\label{Fig: Training_Loss}

\end{figure}

The proposed network is trained and evaluated using KITTI benchmarks\cite{fritsch2013KITTI_road}.
In KITTI dataset, there are 289 training images and 290 testing images
for road detection. The training images have sizes range from $370\times1224$
to $375\times1242$ along with a binary label map presenting the drivable
area. In training session, we augment the data samples by scaling
the original images as well as the ground truth images to $0.5\sim1.0$
of their original resolution and then crop them using a shifting window.
The horizontal shift is 60 pixels and vertical shift is 20 pixels.
Finally, a total of 20,808 samples are generated and separated into
a training set with 20,500 samples and a validating set with 308 samples.
We also add Gaussian noise to the input data with standard deviation
of 0.02\% for additional diversity. Mean absolute error (MAE) of the
boundary location in each column is selected as the loss function.\textcolor{red}{{}
}Adam\cite{Kingma2015} is an gradient descent based optimizer that
adjusts learning rate on each neuron based on the estimation of lower-order
moments of the gradients. We choose Adam as the optimizer because
it converges quickly at the beginning and slows down near convergence.
Input batch size is set to 125, learning rate is fixed at 1e-4. Figure
\ref{Fig: Training_Loss} shows the error loss of the validation data
after each training epoch. After a total 80 epochs training we get
0.0185 MAE on validation data. 

We evaluated the trained network on KITTI\cite{fritsch2013KITTI_road}
test bench. There are two main metrics to evaluate the road segmentation:
F1-score and average precision(AP). The metrics are calculated as
in (\ref{eq: metrics1}-\ref{eq: metrics4}), where $TP$, $TN$,
$FP$, $FN$ denote true positive, true negative, false positive and
false negative. 

\begin{equation}
Precision=\frac{TP}{TP+FP}\label{eq: metrics1}
\end{equation}

\begin{equation}
Recall=\frac{TP}{TP+FN}\label{eq: metrics2}
\end{equation}

\begin{equation}
F1\operatorname{-}score=\frac{2Precision\cdot Recall}{Precision+Recall}\label{eq: metrics3}
\end{equation}

\begin{equation}
AP=\frac{TP+TN}{TP+FP+TN+FN}\label{eq: metrics4}
\end{equation}

The evaluation results obtained from KITTI are F1-score of 86.91\%
and AP of 81.11\%, which is comparable to the state-of-the-art methods
reported so far. In addition, our solution has lower false positive
rate of 4.39\%, which is safe for autonomous vehicle. In Table \ref{Table: Evaluation result}
our work is compared with related solutions listed on KITTI road detection
test bench. It shows that our work has similar F1-score and average
precision with other works but has higher precision and lower false
positive rate. More importantly, the proposed network has much fewer
parameters to train and significantly less floating-point operations.
Our proposed method of road segmentation can achieve real-time speed
at 50 frames per second, when tested on an NVidia GTX 950M CPU with
moderate processing power. We claim that the proposed solution is
among the fastest in KITTI road detection test bench.

\begin{figure}
\begin{tabular}{c}
\includegraphics[width=0.95\columnwidth]{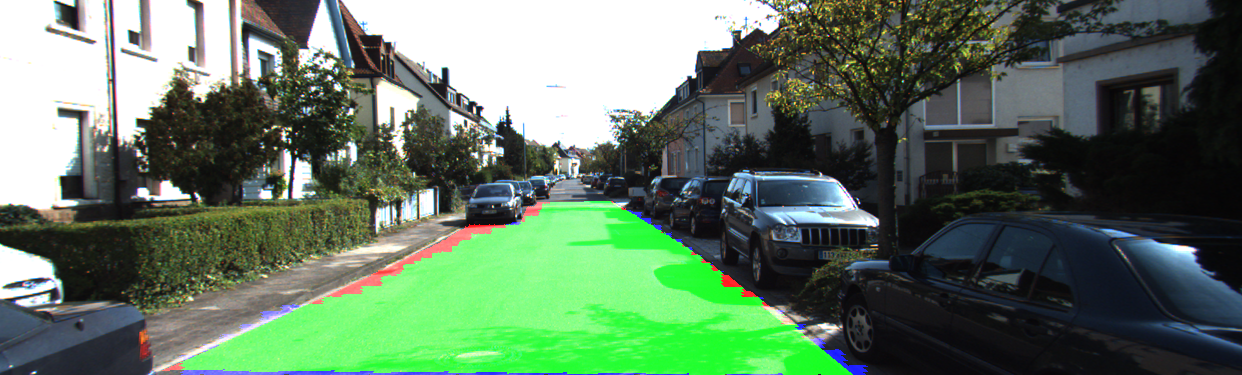}\tabularnewline
\includegraphics[width=0.95\columnwidth]{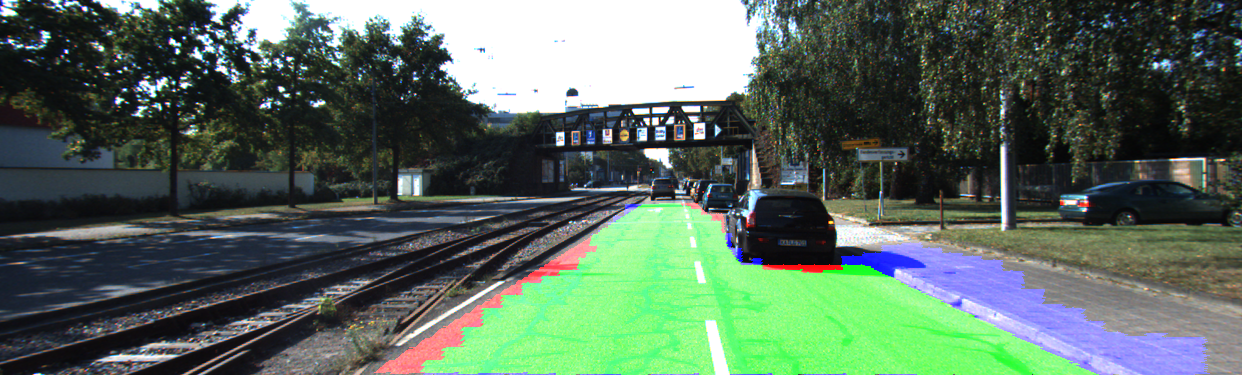}\tabularnewline
\end{tabular}

\caption{A typical road segmentation result from the proposed CNN-RNN }

\label{Fig: Typical Result}
\end{figure}

Figure \ref{Fig: Typical Result} shows the typical result of our
proposed road detector\footnote{FLOPs are estimated from the published results of the neural networks}.
Green pixels are true positives, while blue pixels are false positives
and red pixels are false negatives. It can be seen that the majority
of road surface are detected, and obstacles such as vehicles and railways
are separated to avoid collisions. False negatives usually happen
at road/vehicle and road/sidewalk boundaries, which mostly acceptable
for automated driving. But the false positives on the sidewalks require
further improvement. 

\begin{table*}
\begin{tabular}{|c|c|c|c|c|c|c|c|c|c|}
\hline 
Method &
F1 &
AP &
PRE &
REC &
FPR &
FNR &
Para &
FLOPs &
Runtime\tabularnewline
\hline 
\hline 
ours &
86.91 \% &
81.11\% &
91.24\% &
82.97\% &
4.39\% &
17.03\% &
2.79M &
1.97G &
20ms\tabularnewline
\hline 
StixelNet\cite{levi2015stixelnet} &
89.12 \% &
81.23\% &
85.80\% &
92.71\% &
8.45\% &
7.29\% &
6.82M &
43.0G &
1s\tabularnewline
\hline 
MAP\cite{laddha2016map} &
87.80 \% &
89.96\% &
86.01\% &
89.66\% &
8.04\% &
10.34\% &
457.43M &
7.15G &
280ms\tabularnewline
\hline 
\end{tabular}

\caption{Comparison between networks on the KITTI\cite{fritsch2013KITTI_road}
road segmentation challenge: F1-score (F1), average precision (AP),
precision (PRE), recall (REC), false positive rate (FPR), false negative
rate (FNR), number of parameters (Para), floating operations (FL-OPs)
and runtime}
\label{Table: Evaluation result}
\end{table*}

\section{Conclusion \label{sec:Conclusion}}

In this paper, we present a neural network based solution for road
segmentation that can achieve real-time processing speed. The CNN-RNN
network mainly consists of a light-weighted local feature encoder
and a recurrent neural network to process context information, which
significantly reduces the floating-point operations and the memory
usage. We train the network with KITTI road training database and
evaluate on its test bench. The test result shows that our algorithm
can achieve 86.91\% F1-score and 81.11\% average precision. However,
the image-based road segmentation is still subjected to light conditions.
Shadows, blurs and confusing colors are the main cause of false positives
and false negatives. In our future work, multiple sensors including
cameras, LiDARs and IMUs will be fused to further improve the road
detector performance.

\bibliographystyle{IEEEtran}
\bibliography{13D__Downloads_IV2018_0Reference}

\end{document}